%% file: main.tex
\documentclass[a4paper,conference]{IEEEtran}
\usepackage{cite}
\usepackage{diagbox}
\usepackage{subfig}
\usepackage{setspace}
\usepackage{float}
\usepackage{hyperref}
\usepackage{verbatim}
\usepackage{amsmath}
\usepackage{amssymb}
\usepackage{booktabs}
\usepackage{multirow}
\usepackage{algorithm}
\usepackage{algorithmic}
\usepackage{arydshln}
\usepackage{caption} 
\usepackage{lipsum}
\usepackage{booktabs}
\usepackage{color}
\usepackage{amsfonts}
\usepackage{array,hhline}
\usepackage{wasysym}
\usepackage{graphicx}

\ifCLASSINFOpdf

\else

\fi

\newcommand{\ie}{\textit{i.e.}}
\newcommand{\eg}{\textit{e.g.}}

\def\name{\text{WILD}}
\begin{document}

\title{Removing Backdoor-Based Watermarks in Neural Networks with Limited Data}

\author{
\IEEEauthorblockN{Xuankai Liu}
\IEEEauthorblockA{
Tsinghua University\\
liuxk18@mails.tsinghua.edu.cn}
\and
\IEEEauthorblockN{Fengting Li}
\IEEEauthorblockA{Tsinghua University\\
lft18@mails.tsinghua.edu.cn}
\and
\IEEEauthorblockN{Bihan Wen}
\IEEEauthorblockA{Nanyang Technological University\\
bihan.wen@ntu.edu.sg}
\and
\IEEEauthorblockN{Qi Li}
\IEEEauthorblockA{Tsinghua University\\
qli01@tsinghua.edu.cn}}

\maketitle

\begin{abstract}
Deep neural networks have been widely applied and achieved great success in various fields. As training deep models usually consumes massive data and computational resources, 
trading the trained deep models is highly-demanded and lucrative nowadays. Unfortunately, the naive trading schemes typically involves potential risks related to copyright and trustworthiness issues, \eg, a sold model can be illegally resold to others without further authorization to reap huge profits. To tackle this problem, various watermarking techniques are proposed to protect the model intellectual property, amongst which the backdoor-based watermarking is the most commonly-used one. However, the robustness of these watermarking approaches is not well evaluated under realistic settings, such as limited in-distribution data availability and agnostic of watermarking patterns.
In this paper, we benchmark the robustness of watermarking, and propose a novel backdoor-based watermark removal framework using limited data, dubbed \name{}.
The proposed \name{} removes the watermarks of deep models with only a small portion of training data, and the output model can perform the same as models trained from scratch without watermarks injected.
In particular, a novel data augmentation method is utilized to mimic the behavior of watermark triggers. Combining with the distribution alignment between the normal and perturbed (\eg, occluded) data in the feature space, our approach generalizes well on all typical types of trigger contents.
The experimental results demonstrate that our approach can effectively remove the watermarks without compromising the deep model performance for the original task with the limited access to training data.

\end{abstract}

%
\IEEEpeerreviewmaketitle

\begin{figure*}[t]
	\centering
	\subfloat[]{
		\begin{minipage}[t]{0.25\linewidth}
			\centering
			\includegraphics[width=1.2in]{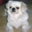}
		\end{minipage}%
	}%
	\subfloat[]{
		\begin{minipage}[t]{0.25\linewidth}
			\centering
			\includegraphics[width=1.2in]{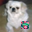}
		\end{minipage}%
	}%
	\subfloat[]{
	\begin{minipage}[t]{0.25\linewidth}
		\centering
		\includegraphics[width=1.2in]{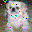}
	\end{minipage}%
	}%
	\subfloat[]{
	\begin{minipage}[t]{0.25\linewidth}
		\centering
		\includegraphics[width=1.2in]{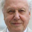}
	\end{minipage}%
	}%
	\centering
	\vspace{0.1in}
	\caption{Various kinds of backdoor-based watermarks. (a) shows an unperturbed image of a dog, while (b), (c) and (d) demonstrate three different watermarks, i.e., content-based, noise-based, and unrelated watermarks. The labels of these watermarked images are tampered with into a wrong label, such as a truck.}
	\label{fig:watermarks}
\end{figure*}

\section{Introduction}

With the rapid development of deep learning, deep neural networks (DNNs) have been widely applied in many fields, such as computer vision and speech recognition~\cite{lawrence1997face,amodei2016deep,toshev2014deeppose,bahdanau2014neural}. In practice, training of DNNs is usually expensive, \ie, consuming a huge amount of high-quality data and computational resource.
Therefore, transactions on well-trained models open a new market of Machine Learning as a Service (MLaaS)~\cite{ribeiro2015mlaas,li2017scaling,tafti2017machine}, which has been developed into a profitable industry. 
However, protection of the model copyright becomes a challenging problem, \ie, deep models can be redistributed for illegal profits once released, which could cause significant economic damage to the legitimate owners.

Various watermarking schemes have been proposed for model protection~\cite{rouhanideepsigns,zhang2018protecting,le2019adversarial,uchida2017embedding}. 
Amongst them, the backdoor-based watermarking is currently the most trendy in deployment, \eg, companies such as IBM have already planed to apply the backdoor-based watermarking technique in their authorized neural network models~\cite{adi2018turning,zhang2018protecting}. 
Watermarks can be infused into the deep models by injecting poisonous data  during  training.
These poisonous data contain watermark patterns, with original labels modified to  wrong labels. The owner of the model can claim the ownership by verifying whether the outputs are pre-defined wrong labels by the presence of watermarks. In this way, the intellectual property of the model can be effectively protected.

Recent works proposed to remove watermarks from deep models, by either fine-tuning or training regularization approaches. 
The first approach, by leveraging the auxiliary unlabeled data to fune-tune the watermarked model, can decrease the amount of labeled training data needed for effective watermark removal~\cite{chen2019leveraging}.
However, collecting such a large amount of additional unlabeled data  is  a huge burden for adversaries in practice. 
Besides, REFIT~\cite{chen2019refit:} demonstrated that using a large learning rate during fine-tuning can remove the watermark, but it could cause great damage to test accuracy. 
Though REFIT applied elastic weight consolidation (EWC)~\cite{kirkpatrick2017overcoming} to avoid the degradation  of the model performance, such method heavily relies on a carefully-designed learning rate schedule which is hard for generalization.
Alternatively, the training regularization approach aims to prevent overfitting to the backdoor-based watermarks~\cite{shafieinejad2019robustness}, \ie, the watermarking samples are regarded as the ``outliers'' to the normal data distribution, which the watermarked model overfits to, thus, the $\ell_2$-based regularizer is imposed in training to remove watermarks.
However, such approach assumes that the adversaries have the same amount of in-distribution data as  training set, which is usually invalid in practice.
In other words, with the access to the adversaries which include the entire training set, or the same amount of  in-distribution data as the training set, one can easily reproduce a new model from scratch, instead of only removing the existing model watermarks. 
In practice, there is only very limited access to the original training set (\eg, a small subset of the in-distribution data), which requires the watermark removal algorithm to work under the weakly supervised setting.

In this work, we propose a novel backdoor-based watermark removal framework with very limited access to the in-distribution data, dubbed \name{}.
There are two main features and objectives of the proposed WILD scheme:
(i) a novel data augmentation method is proposed to enhance the model robustness against the watermark patterns, which can be treated as occlusions on top of the normal image inputs. 
(ii) the processed model by WILD needs to maintain the similar effectiveness of analyzing the images with or without watermarks, for which we minimize the distribution gap between them in the high-level feature space.
We evaluate the proposed \name{} with extensive experiments over the MNIST and CIFAR-10 datasets. 
The results show that the watermark retention degrades from almost 100\% to below 20\%, while the test accuracy drops
by 0.3\%  on MNIST  and 1.7\% on CIFAR-10 on average when 20\% of the training data is available.
It demonstrates that \name{}  can effectively remove the watermarks using only a small proportion of data, with limited impact on test accuracy.


In summary, we make the following contributions:
\begin{itemize}

\item We propose \name{} for watermark removal against backdoor-based watermark techniques. Compared with prior work, only a small proportion of data are demanded in \name{} rather than the entire training set. Besides, the proposed WILD does not require any additional unlabeled data or carefully-designed learning rate schedule.

\item We invalidate the model watermarks from the perspective of improving the robustness of the model. Specifically, we incorporate two techniques including data augmentation and the optimization of distribution distance in high-level feature space.

\item We evaluate our approach on MNIST and CIFAR-10 datasets respectively. The experimental results demonstrate that our approach can effectively remove these types of watermarks with limited impact on the performance on original task.


\end{itemize}

\begin{figure}[t]
	\centering 
	\includegraphics[width=1.05\linewidth]{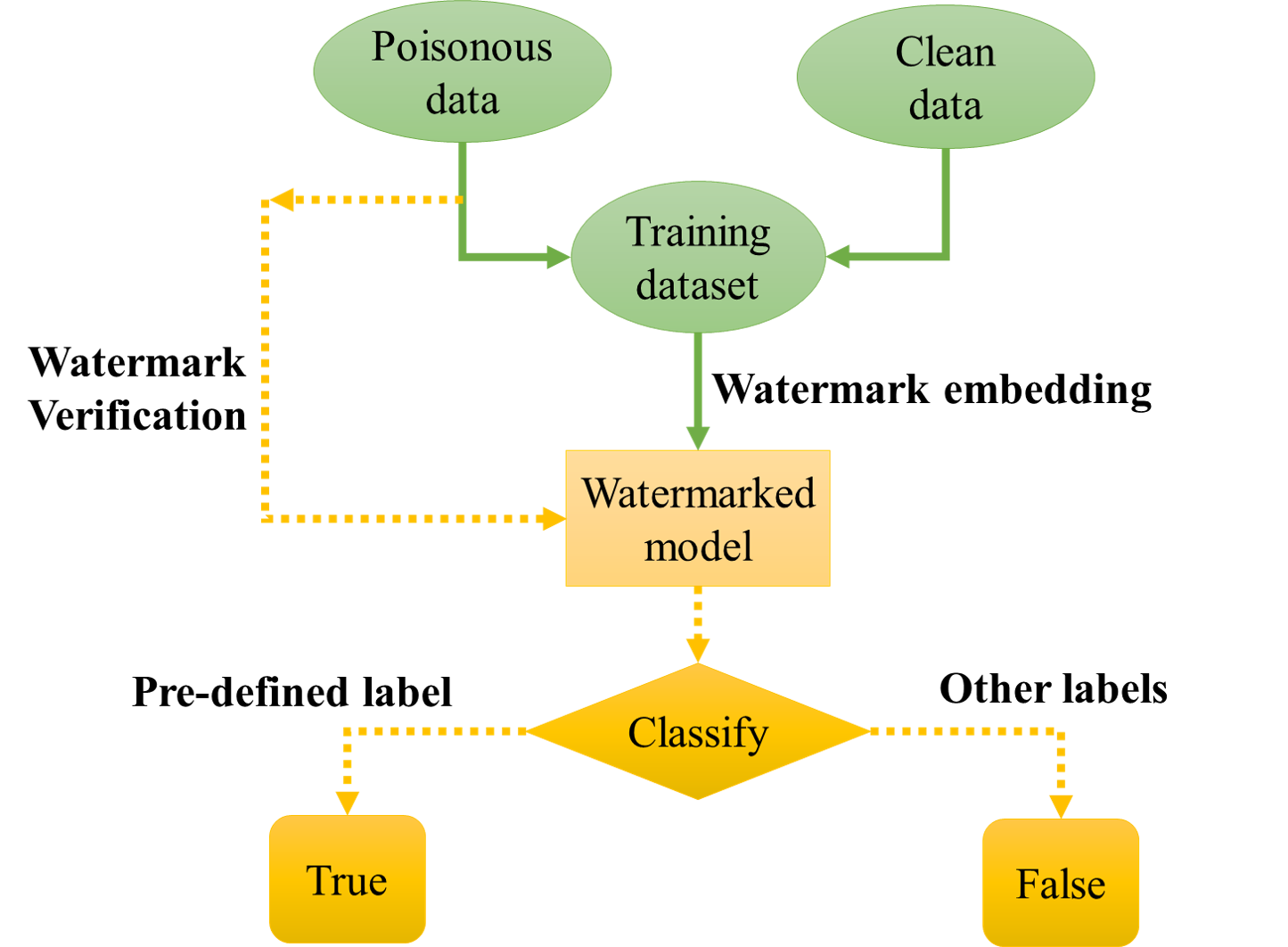}
	\vspace{0.1in}
	\caption{An illustration of watermark embedding and verification process for backdoor-based watermarks. Poisonous data is added in the training dataset to embed the watermark into the model during training. Besides, these poisonous data can be used to extract the watermark by verifying the outputs during the testing phase.}
	\label{fig:intuition}
\end{figure}

\section{Background}

\subsection{Neural Network Backdoors}


A DNN backdoor is a hidden pattern trained into the neural network model. The model could misbehave by the presence of trigger pattern. There are two  methods of implementing backdoor attack. One is poisoning attack~\cite{chen2017targeted,wang2019neural} and the other is trojaned attack~\cite{liu2018trojaning}. Poisoning attack is commonly used to achieve backdoor attack. Adversaries generate poisonous data by adding the  trigger pattern to clean data, then tamper with their original labels to wrong labels. These poisonous data participate in the training of the model. In this way, the model memorizes the trigger pattern and could misclassify the poisonous inputs into pre-defined labels. Trojaned attack is another way to inject malicious behaviors into the model but has no need to tamper with the original training process. It generate the trojan trigger that can  induce evident activation in some neurons inside the model, then retrain the layers between the activated neurons and the output layer to respond to a specific output. In this way, the malicious behavior is implanted. Specifically, Gu et al.\cite{gu2017badnets} proposed BadNets, which aim to create a maliciously trained network. It has state-of-the-art performance on benign inputs but behave abnormally on malicious samples. This work not only demonstrate the backdoor attack on the MNIST dataset~\cite{lecun1998gradient}, but also use a simple sticker to deceive the road sign classifier which is widely applied in autonomous driving. Chen et al.\cite{chen2017targeted} studied backdoor poisoning attacks and they achieved backdoor attacks by injecting poisoning data into the training set. Liu et al.\cite{liu2017trojaning} demonstrated that they could achieve trojaning attack without access to the dataset that are used to train the model. They first reverse engineer model inputs, and then use them to retrain the model to inject malicious behaviors to the model.

\subsection{Backdoor Defenses}
Baracaldo et al.~\cite{baracaldo2017mitigating} proposed the defense strategy that uses data provenance to filter untrusted data points and prevent poisoning attacks. Liu et al.~\cite{liu2018fine} combined fine-tuning and pruning to disable backdoor attacks. They observe the different behavior of dormant neurons activated by benign and malicious inputs respectively. Wang et al.~\cite{wang2019neural} identified whether there is an input trigger that would produce misclassified results, then reversed engineer triggers which was formulated as a multi-objective optimization task. Chen et al.~\cite{chen2018detecting} proposed the Activation Clustering (AC) method for detecting poisonous training data. They found that the difference between benign and poisonous inputs is evident in the network activations which was leveraged to set them apart. Gao et al.~\cite{gao2019strip} observed the randomness of predicted classes after perturbing the input by superimposing various image patterns, and a low entropy in predicted classes implied the presence of a malicious trojan trigger due to the characteristic of a trojaned input. Chou et al.~\cite{chou2018sentinet} leveraged techniques from model interpretability to discover the suspicious region that may contain trojan trigger, and then observed the model outputs after overlaying the suspected region on a set of benign test images.

\subsection{Backdoor-based Watermark in DNNs}
Although backdoor attacks are harmful to neural networks, 
recent works demonstrate that backdoor techniques can be applied as watermarks to protect the intellectual property of the model~\cite{zhang2018protecting,adi2018turning}. 
There are three typical backdoor-based watermarking schemes~\cite{zhang2018protecting}. The first one is content-based watermark. In detail, we blend 
certain meaningful content on top of the clean images as watermarks such as a small piece of sticker. The second type of watermark scheme is based on noise patterns. Different from meaningful content, we add specific noise (e.g.,  Gaussian noise) on the  images. Another way is to utilize some unrelated images that come from totally different domains. For example, if our original task is to train a model to classify different kinds of animals, we may choose face images as watermarks. Figure~\ref{fig:watermarks} shows the examples of these three types of watermarks. The original image is  a image of dog, and we  assign an "truck" label to the watermarked images. 
These watermarked images participate in the training of the watermarked model. In this way, we can claim the ownership of the model if the model recognizes these images as the  pre-defined label (truck). The processes of watermark embedding and verification are illustrated in Figure~\ref{fig:intuition}.


\subsection{Watermark Removal}
Uchida et al.~\cite{uchida2017embedding} proposed the first watermark method in DNNs, leveraging a parameter regularizer to embed the watermark in model parameters. However, Wang et al.~\cite{wang2019attacks} demonstrated that this type of watermark technique modifies the statistical distribution of the model, and this modification can be used not only detect the presence of a watermark but even derive its embedding length and use this information to remove the watermark by overwriting it. Shafieinejad et al.~\cite{shafieinejad2019robustness} focused on backdoor-based watermarking. Their watermark removal relied on unlabeled public data and black-box access to the classification label. Regularization and fine-tuning are combined to remove the watermark and avoid overfitting under a white-box setting.
Similarly, Chen et al.~\cite{chen2019leveraging} demonstrated that leveraging unlabeled auxiliary data significantly decreases the amount of labeled training data needed for effective watermark removal. However, it is hard for adversaries to collect a large amount of unlabeled data, which are in the same distribution as the original training data. Once these unlabeled data are not from the same distribution, then the process of fine-tuning could impact the performance of the model on the original task. Although watermark techniques allow the legitimate owner to detect copyright infringements of his model, Hitaj et al.~\cite{hitaj2018have} demonstrated that a malicious adversary could evade the verification by the legitimate owners, thus avoiding the detection of model theft.

\begin{figure}[t]
	\centering
	\subfloat[]{
		\begin{minipage}[t]{0.33\linewidth}
			\centering
			\includegraphics[width=1.1in]{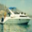}
		\end{minipage}%
	}%
	\subfloat[]{
		\begin{minipage}[t]{0.33\linewidth}
			\centering
			\includegraphics[width=1.1in]{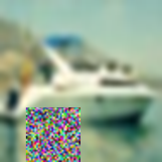}
		\end{minipage}%
	}%
		\subfloat[]{
		\begin{minipage}[t]{0.33\linewidth}
			\centering
			\includegraphics[width=1.1in]{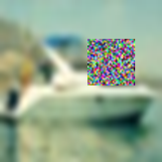}
		\end{minipage}%
	}%

	\subfloat[]{
		\begin{minipage}[t]{0.33\linewidth}
			\centering
			\includegraphics[width=1.1in]{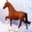}
		\end{minipage}%
	}%
	\subfloat[]{
		\begin{minipage}[t]{0.33\linewidth}
			\centering
			\includegraphics[width=1.1in]{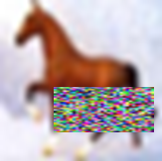}
		\end{minipage}%
	}%
		\subfloat[]{
		\begin{minipage}[t]{0.33\linewidth}
			\centering
			\includegraphics[width=1.1in]{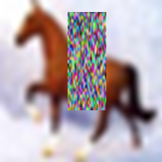}
		\end{minipage}%
	}%

	\centering
	\vspace{0.1in}
	\caption{An illustration of augmented images using Random Erasing. The leftmost column is the normal data. (b) (c) and (e) (f) are occluded images of (a) and (d), respectively. Erasing random regions in training data can make the model robust to occlusion.}
	\label{fig:noise}
\end{figure}

\begin{figure*}[t]
	\centering 
	\includegraphics[width=0.96\linewidth]{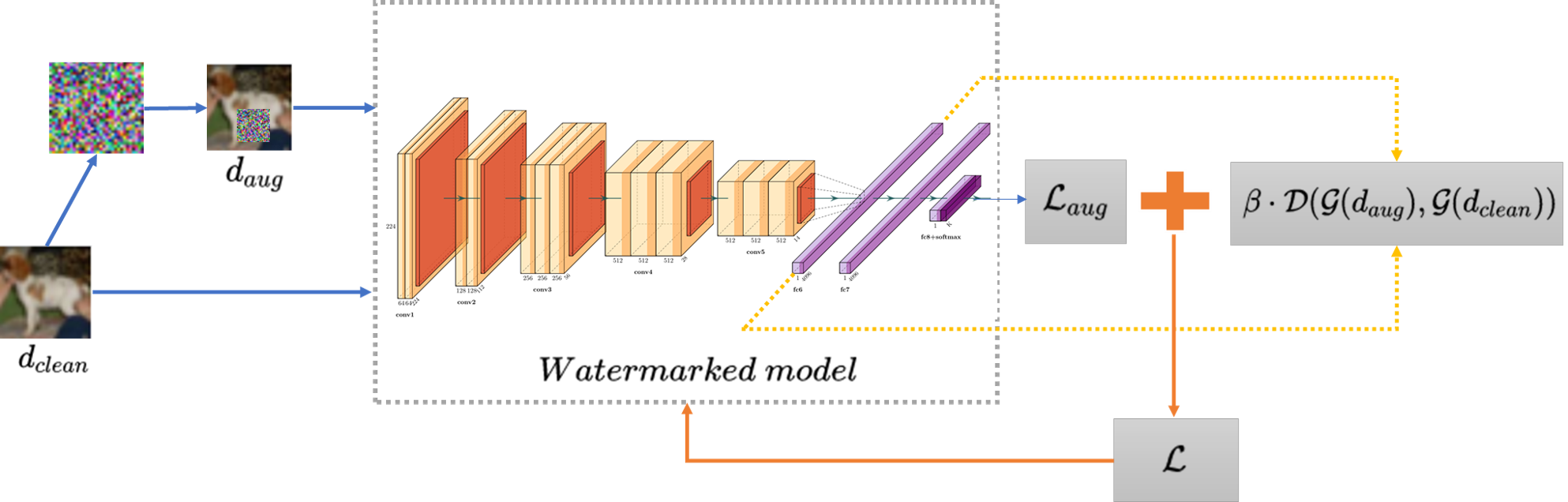}
	\vspace{0.1in}
	\caption{An overview of WILD. The proposed WILD incorporates two schemes, i.e, data augmentation, and alignment of feature distribution.}
	\label{fig:architecture1}
\end{figure*}

\section{System Design}



In this section we present our framework to remove watermarks in neural networks, by introducing a strawman approach in Section~\ref{sec:strawnman}, followed by a novel distribution alignment approach that fully leverages feature spaces to effectively remove watermarks in Section~\ref{sec:our approach}.

\subsection{A Strawman Approach}
\label{sec:strawnman}

It is challenging to identify watermark triggers in images, \ie, exhaustive search of the trigger pattern amongst all pixels is NP-hard.
Besides, heuristically reverse-engineering the trigger pattern is inaccurate and time-consuming~\cite{wang2019neural}, which may further impair the watermark removal ability.
In practice, adversarial fine-tuning techniques have been widely discussed for watermark removal, which is more tractable and effective.

The types of watermark patterns that can be injected into a neural network model are diversified. Thus, we aim to develop an effective approach that can correctly recognize all these potential watermarked data. Despite the difference in the shape or location, the common pattern of watermarks is that they occlude a part of the clean inputs. Therefore, we can turn the problem of identifying watermarks into prompting the model to correctly identify the occluded data. 
We utilize Random Erasing~\cite{zhong2020random} to fine-tune the watermarked model robust against occlusion.
Specifically, we randomly select some rectangle regions in the clean data and fill them with random Gaussian noises to augment the data. Examples are shown in Figure~\ref{fig:noise}. However, we find that data augmentation alone is insufficient 
for watermark removal due to the diversity of the watermarks and the strong bias towards injected poisonous data. 

\subsection{Our Distribution Alignment Approach}
\label{sec:our approach}
Now we consider a high-level feature space and improve the effectiveness of backdoor-based watermark removal from the perspective of feature distribution. 
The intuition behind is that the injected watermarks in neural networks form strongly correlated paths from the input layer to the output layer of the model. Also,~\cite{adi2018turning} shows that retraining the fully-connected layers alone contributes little to removing the watermarking effect, which side proves that these paths are dominated by some special convolutional kernels generating large activation values with the presence of a watermark pattern. As these activations largely surpass the activation of the original object, the model outputs the incorrect label desired by the model owner. Thus, we can mitigate the impact of watermarks by weakening the effectiveness of these paths to make the benign object back in control. A good start point that sees all these special kernels is in the high-level feature space just after the final convolutional layer.


Thus, when data points with watermark pattern (\ie $d'$) are similar to the clean data $d_{clean}$ in the input space, we hope that the distribution of high-level features $\mathcal{G}(d')$ is also similar to $\mathcal{G}(d_{clean})$. Therefore, we make efforts to minimize the distribution distance between them, and formalize this optimization problem as follows:


\begin{equation}
\begin{array}{cl}
\operatorname{minimize} & \mathcal{D}\left(\mathcal{G}(d_{clean}), \mathcal{G}(d’)\right), \\
\operatorname{such\ that} & d’ \in \mathcal{A}\left(d_{clean}, w, l\right),
\end{array}
\end{equation}
where $\mathcal{A}(d_{clean}, w, l)$ denotes applying watermark trigger $w$ at location $l$ on data point $d$,
and $\mathcal{D}$ is a metric function measuring the distance between the two distributions in high-level feature space, i.e., $\mathcal{G}(d_{clean})$ and $\mathcal{G}(d')$.

To minimize the distributions distance, a straightforward way is to penalize the distribution distance during the process of fine-tuning. Since we have no prior knowledge which form of the watermark is embedded into the model, we cannot directly calculate the distribution distance between watermarked data and clean data. Note that augmented data $d_{aug}$ discussed in Section~\ref{sec:strawnman} are also a  special form of perturbed data compared to clean data, 
so $d_{aug}$ can be regarded as an approximation of $d'$. In this way, our approach incorporates two strategies. On the one hand, we fine-tune the watermarked model using the augmented data $d_{aug}$, and we denote the calculated loss as $\mathcal{L}_{aug}$. On the other hand, we add a penalty item for the distribution distance $\mathcal{D}(\mathcal{G}(d_{clean}), \mathcal{G}(d_{aug}))$. Therefore, the loss function $\mathcal{L}$  in our framework is denoted as follows:    
\begin{equation}
\mathcal{L} = \mathcal{L}_{aug} + \beta \cdot  \mathcal{D}(\mathcal{G}(d_{clean}), \mathcal{G}(d_{aug})),
\end{equation}
where $\beta$ is the penalty coefficient that regulates the strength of the penalty item. The architecture of our approach is shown in Figure~\ref{fig:architecture1}. We first apply Random Erasing for data augmentation to generate the augmented data $d_{aug}$, then $d_{aug}$ and clean data $d_{clean}$ are sent into the watermarked model. Note that $d_{clean}$ are only used to measure the distribution distance $\mathcal{D}(\mathcal{G}(d_{clean}), \mathcal{G}(d_{aug}))$ and the loss calculated by 
the outputs of $d_{clean}$
and their true labels does not participate in the fine-tuning process. The features after the last convolutional layer are utilized  to measure the distribution distance. They  are flatten into a one-dimensional vector, then normalized by a softmax function. As for the metric function $\mathcal{D}$,  two different metric functions are considered separately in our work:

\subsubsection{Cross-entropy} Cross-entropy is commonly used as loss function when optimizing multi-class classification models. It measures the distance between the output probability distribution and the target  distribution. The smaller the value of  cross-entropy, the closer the two distributions are. 
The cross-entropy between two distributions $P$ and $Q$ is as follows:
\begin{equation}
H(P, Q) = – \sum_{x}  P(x) \cdot log(Q(x)),
\end{equation}
where $P$ is the target distribution and $Q$ is the approximation of the target distribution.

\subsubsection{Jensen–Shannon divergence} Jensen–Shannon (JS) divergence is also a widely used metric function to measure the similarity of the two probability distributions. Based on the variation of the Kullback–Leibler (KL) divergence, JS divergence solves the asymmetric problem of the KL divergence, with its value  between 0 and 1. The JS divergence is as follows:
\begin{equation}
JS(P||Q) = \frac{KL(P|| \frac{P+Q}{2}) + KL(Q|| \frac{P+Q}{2})}{2},
\end{equation}
where $KL$ denotes the KL divergence.

To comprehensively assess the performance of our approach, cross-entropy and JS divergence will be applied as metric function $D$ respectively to measure the distribution distance. Apart from explicitly define the distance metrics, we could also use a discriminative network to learn the distribution mapping between clean and augmented data points, thus making the watermarked model to generate indistinguishable high-level features for $d_{clean}$ and $d_{aug}$. All three metrics will be evaluated in the next section.

\begin{figure}[t]
	\centering
	\subfloat[MNIST, content]{
		\begin{minipage}[t]{0.33\linewidth}
			\centering
	\includegraphics[width=1.05in]{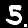}
		\end{minipage}%
	}%
	\subfloat[MNIST, noise]{
	\begin{minipage}[t]{0.33\linewidth}
		\centering
		\includegraphics[width=1.05in]{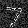}
	\end{minipage}%
	}%
	\subfloat[MNIST, unrelated]{
	\begin{minipage}[t]{0.33\linewidth}
		\centering
		\includegraphics[width=1.05in]{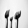}
	\end{minipage}%
	}%

	\subfloat[CIFAR-10, content]{
	\begin{minipage}[t]{0.33\linewidth}
		\centering
\includegraphics[width=1.05in]{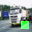}
	\end{minipage}%
}%
	\subfloat[CIFAR-10, noise]{
	\begin{minipage}[t]{0.33\linewidth}
		\centering
		\includegraphics[width=1.05in]{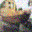}
	\end{minipage}%
	}%
	\subfloat[CIFAR-10, unrelated]{
	\begin{minipage}[t]{0.33\linewidth}
		\centering
		\includegraphics[width=1.05in]{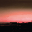}
	\end{minipage}%
	}
	\centering
	\vspace{0.1in}
	\caption{Examples of three different types of watermarked data used in our experiment. The first row shows the watermarked data in MNIST, while the images in the second row are drawn from CIFAR-10.}
	\label{fig:watermarkExamples}
\end{figure}

\begin{figure*}[t]
	\centering
	\subfloat[MNIST, content-based]{
		\begin{minipage}[t]{0.33\linewidth}
			\centering
	\includegraphics[width=2.3in]{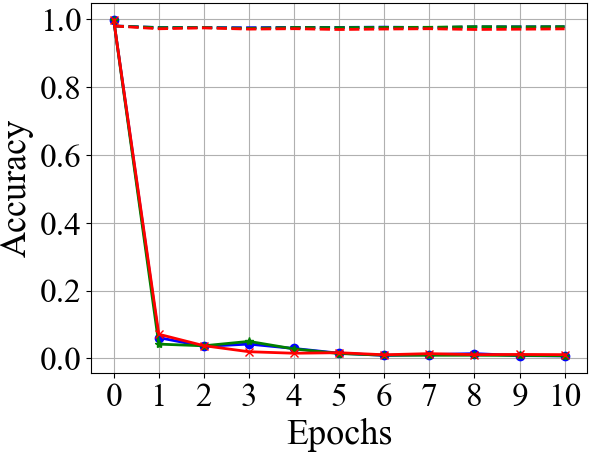}
		\end{minipage}%
	}%
	\subfloat[MNIST, noise-based]{
	\begin{minipage}[t]{0.33\linewidth}
		\centering
		\includegraphics[width=2.3in]{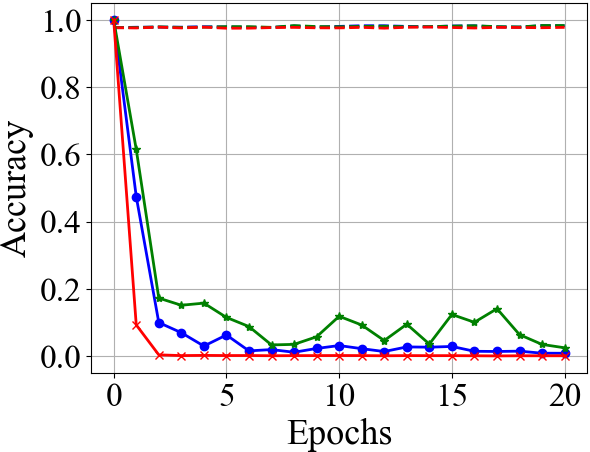}
	\end{minipage}%
	}%
	\subfloat[MNIST, unrelated]{
	\begin{minipage}[t]{0.33\linewidth}
		\centering
		\includegraphics[width=2.3in]{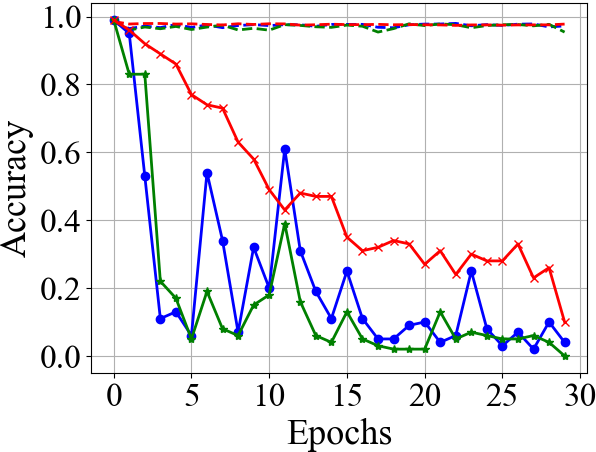}
	\end{minipage}%
	}%
	
		\subfloat[CIFAR-10, content-based]{
		\begin{minipage}[t]{0.33\linewidth}
			\centering
	\includegraphics[width=2.3in]{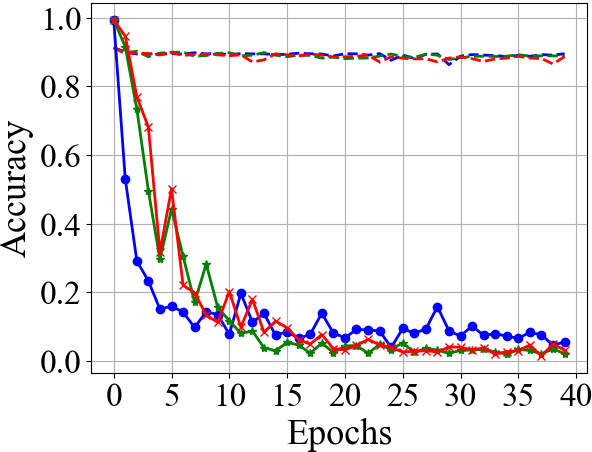}
		\end{minipage}%
	}%
	\subfloat[CIFAR-10, noise-based]{
	\begin{minipage}[t]{0.33\linewidth}
		\centering
		\includegraphics[width=2.3in]{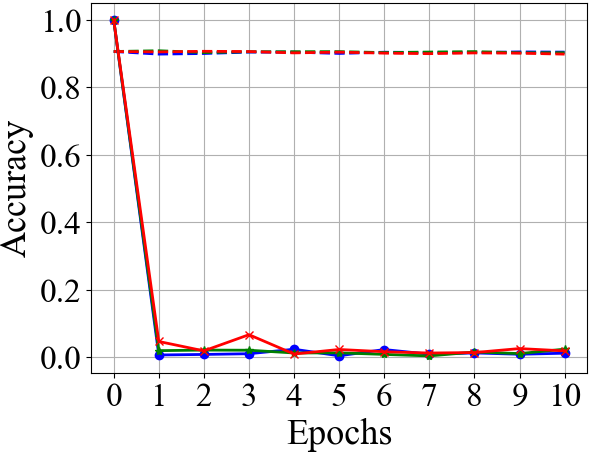}
	\end{minipage}%
	}%
	\subfloat[CIFAR-10, unrelated]{
	\begin{minipage}[t]{0.33\linewidth}
		\centering
		\includegraphics[width=2.3in]{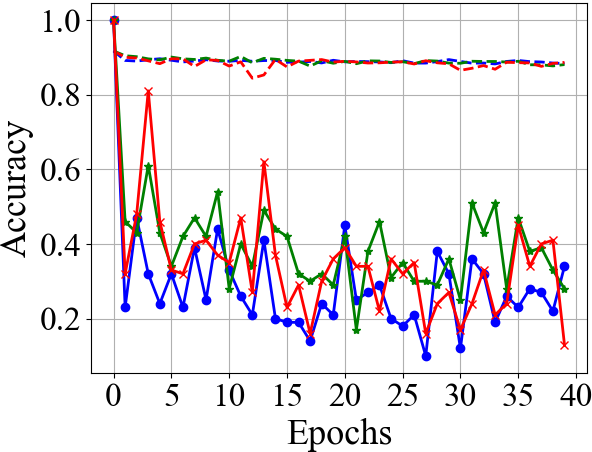}
	\end{minipage}%
	}%
	
		\subfloat{
	\begin{minipage}[t]{1.0\linewidth}
		\centering
		\includegraphics[width=3.5in]{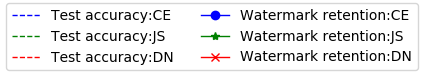}
	\end{minipage}%
	}%
	\centering
	\caption{Test accuracy and watermark retention curves of three approaches. CE, JS and DN  denote cross-entropy, JS divergence and discriminative network approaches respectively.}.
	\label{fig:curves}
\end{figure*}

\section{Experiment}

\subsection{Evaluation}
We evaluate our approach on MNIST and CIFAR-10 datasets. We use LeNet model for MNIST and VGG-16 model for CIFAR-10. $RMSProp$ optimizer with initial learning rate of 0.004 is used to train the watermarked model. For all experiments of watermark removal, $RMSProp$ optimizer is also used to  retrain the watermarked model with the initial learning rate set to be 0.001. The detailed experiment settings are as follows:

\begin{itemize}

\item  To embed pattern-based watermarks, square trigger at the bottom-right corner is used following the setting in~\cite{gao2019strip:}. Besides, we add Gaussian noise to inject noise-based watermarks using the setting in~\cite{zhang2018protecting}.
As for  unrelated watermarks, we follow the setting in~\cite{adi2018turning} and leverage the 100 images in trigger set which are totally unrelated to our training data. Some examples of watermarked data in our experiment are shown in Figure~\ref{fig:watermarkExamples}. 
The watermarked models are trained to misclassify data points applied with watermark trigger into label 0 (i.e, label airplane in CIFAR-10 and digital 0 in MNIST respectively). We ensure that all the watermark accuracy (using watermark inputs to fool the watermarked model) are approximately  100\%. 

\item According to our premise settings, we only have access to limited data. In our experiment, we evaluate our approach using 10\%, 20\% and 40\% of training data. These clean data are augmented ten times using Random Erasing, that is, for one image, ten rectangle regions in the images are randomly chosen and assigned to random noise. Besides, the adversaries who aim to remove the watermarks can sweep a 
wide range of hyper-paramete to select the best configuration. The setting of the penalty coefficient $\beta$ in our experiment is  shown in Table~\ref{tab:parameter}.

\item Instead of penalizing the distance between the two distributions directly,
here we introduce a discriminative network composed of four-layer fully connected layers which serves as a binary classifier. The  features after the last convolutional layer in watermarked model are sent into the discriminative network. It is trained to determine whether the features come from $d_{clean}$ or $d_{aug}$, with the target label of $d_{aug}$  set  to 0 and $d_{clean}$ set to 1. Besides the training loss $\mathcal{L}_{aug}$, we add a loss item  calculated by the outputs 
of $d_{aug}$  and label 1. In this way, we push the feature distribution $\mathcal{G}(d_{aug})$ close to $\mathcal{G}(d_{clean)}$, and it is consistent with our goal discussed in Section~\ref{sec:our approach}.
In the following experiment, we compare the performance of watermark removal between our approach (adding  penalty for distribution distance based on cross-entropy and JS divergence) and discriminative network approach. We denote cross-entropy, JS divergence and discriminative network as CE, JS and DN respectively.

\item Besides, we also compare our approach with REFIT~\cite{chen2019refit:},  which is the state-of-the-art watermark removal framework. To ensure the consistency of the experimental setting, we continue to use $RMSProp$ optimizer. As discussed before,  REFIT relies on the carefully-designed learning rate schedule, therefore, we evaluate the performance of REFIT using different learning rates.

\end{itemize}

\begin{table}[t] 
\renewcommand\arraystretch{1.6}
    \centering
    \fontsize{9}{11}\selectfont    
    \caption{The setting of the penalty coefficient $\beta$ for different datasets in our experiments.}
    \vspace{0.05in}
    \setlength{\tabcolsep}{3.6mm}
\begin{tabular}{c|cc}
\toprule
\diagbox{\bf{Dataset}}{\bf{Metric}} & {\bf{Cross-entropy}} & {\bf{JS Divergence}} \\
\hline
\bf{MNIST} & 0.1 & 1  \\
\hline
\bf{CIFAR-10} &  2 & 10 \\   %

    \bottomrule
\end{tabular}\vspace{0cm}
    \label{tab:parameter}
\end{table}

\begin{table*}[t]\Huge
\renewcommand\arraystretch{1.7}
\caption{Detailed results of test accuracy and watermark retention before and after watermark removal on MNIST.}
\vspace{0.05in}
\label{table:mnist}
\resizebox{0.98\textwidth}{!}{

\begin{tabular}{c|c|c|c|c|c|c|c|c|c}

\toprule[1.5pt]
\multicolumn{1}{c|}{\multirow{3}{*}{\textbf{\begin{tabular}[c]{@{}c@{}}Watermark\\ Type \end{tabular}}}} &\multicolumn{1}{c|}{\multirow{3}{*}{ \textbf{\begin{tabular}[c]{@{}c@{}}Original\\ Watermark Retention\end{tabular}} }}&\multicolumn{1}{c|}{\multirow{3}{*}{\textbf{\begin{tabular}[c]{@{}c@{}}Original\\ Test Accuracy\end{tabular}}}}&\multicolumn{1}{c|}{\multirow{3}{*}{\textbf{\begin{tabular}[c]{@{}c@{}}Percentage\\ of data\end{tabular}}}} &\multicolumn{6}{c}{\textbf{Test Accuracy / Watermark Retention after 
Watermark Removal}} \\ \cline{5-10}
& & & & \multicolumn{3}{c|}{\textbf{REFIT}} &

\multicolumn{3}{c}{\textbf{Ours}} \\ \cline{5-10}
& & & & \textbf{lr = 0.001} & \textbf{lr = 0.003} &\textbf{lr = 0.005} &\textbf{CE}& \textbf{JS}& \textbf{DN} \\\midrule[1.5pt]
\multicolumn{1}{c|}{\multirow{3}{*}{\textbf{Content-based}}} &\multicolumn{1}{c|}{\multirow{3}{*}{99.86\%}}&\multicolumn{1}{c|}{\multirow{3}{*}{98.08\%}}& 10\% & 97.79\% / 26.33\% &98.35\% / 6.55\% &97.68\% / 3.27\% &97.64\% / 1.11\%& 97.57\% / 0.92\% & 97.03\% / 0.62\%  \\
\multicolumn{1}{c|}{} & \multicolumn{1}{c|}{} & \multicolumn{1}{c|}{} & 20\% & 98.81\% / 17.71\% &98.48\% / 5.46\% &97.97\% / 2.63\% &97.74\% / 0.82\% &97.68\% / 0.86\% &97.17\% / 1.16\%  \\
\multicolumn{1}{c|}{} & \multicolumn{1}{c|}{} & \multicolumn{1}{c|}{} & 40\% & 98.87\% / 7.78\% &98.67\% / 3.17\% &98.34\% / 0.31\% & 97.79\% / 0.53\% &97.75\% / 0.92\% &97.24\% / 0.43\%  \\\hline

\multicolumn{1}{c|}{\multirow{3}{*}{\textbf{Noise-based}}} &\multicolumn{1}{c|}{\multirow{3}{*}{100.00\%}}&\multicolumn{1}{c|}{\multirow{3}{*}{97.64\%}}& 10\% &98.73\% / 99.11\% &98.14\% / 98.35\% &97.86\%  / 56.83\% &97.75\% / 4.17\% & 97.82\%  / 6.26\% & 97.44\% / 0.65\% \\
\multicolumn{1}{c|}{} & \multicolumn{1}{c|}{} & \multicolumn{1}{c|}{} & 20\% & 98.93\% / 98.98\% &98.62\% / 96.74\% &98.39\% / 48.52\% & 98.07\% / 0.82\% & 98.02\% / 3.43\% &97.68\% / 0.13\%   \\
\multicolumn{1}{c|}{} & \multicolumn{1}{c|}{} & \multicolumn{1}{c|}{} & 40\% & 99.03\% / 76.85\% &98.73\% / 70.33\% &98.69\% / 46.01\% & 98.14\% / 2.32\% &98.13\% / 4.92\% &97.73\% / 0.17\%  \\\hline

\multicolumn{1}{c|}{\multirow{3}{*}{\textbf{Unrelated}}} &\multicolumn{1}{c|}{\multirow{3}{*}{99.00\%}}&\multicolumn{1}{c|}{\multirow{3}{*}{98.53\%}}& 10\% & 99.03\% / 89.00\% &98.92\% / 84.00\% & 98.77\% / 69.00\% & 97.37\% / 11.00\% & 97.17\% / 15.00\%  &97.33\%  / 9.00\% \\
\multicolumn{1}{c|}{} & \multicolumn{1}{c|}{} & \multicolumn{1}{c|}{} & 20\% &99.05\% / 86.00\% &98.89\% / 79.00\% &98.68\% / 58.00\% & 97.81\% / 10.00\% &97.60\% / 7.00\%& 97.56\% / 14.00\%  \\
\multicolumn{1}{c|}{} & \multicolumn{1}{c|}{} & \multicolumn{1}{c|}{} & 40\% & 99.22\% / 78.00\% &98.94\% / 55.00\% &98.85\% / 35.00\% & 97.95\% / 9.00\% &97.82\% / 6.00\% &97.71\% / 11.00\%  \\

\bottomrule[1.5pt]
\end{tabular}
}
\end{table*}

\begin{table*}[t]\Huge
\renewcommand\arraystretch{1.7}
\caption{Detailed results of test accuracy and watermark retention before and after watermark removal on CIFAR-10.}
\vspace{0.05in}
\label{table:cifar}
\resizebox{0.98\textwidth}{!}{

\begin{tabular}{c|c|c|c|c|c|c|c|c|c}
\toprule[1.5pt]
\multicolumn{1}{c|}{\multirow{3}{*}{\textbf{\begin{tabular}[c]{@{}c@{}}Watermark\\ Type \end{tabular}}}} &\multicolumn{1}{c|}{\multirow{3}{*}{ \textbf{\begin{tabular}[c]{@{}c@{}}Original\\ Watermark Retention\end{tabular}} }}&\multicolumn{1}{c|}{\multirow{3}{*}{\textbf{\begin{tabular}[c]{@{}c@{}}Original\\ Test Accuracy\end{tabular}}}}&\multicolumn{1}{c|}{\multirow{3}{*}{\textbf{\begin{tabular}[c]{@{}c@{}}Percentage\\ of data\end{tabular}}}} &\multicolumn{6}{c}{\textbf{Test Accuracy / Watermark Retention after 
Watermark Removal}} \\ \cline{5-10}
& & & & \multicolumn{3}{c|}{\textbf{REFIT}} &

\multicolumn{3}{c}{\textbf{Ours}} \\ \cline{5-10}
& & & & \textbf{lr = 0.001} & \textbf{lr = 0.003} &\textbf{lr = 0.005} &\textbf{CE}& \textbf{JS}& \textbf{DN} \\\midrule[1.5pt]
\multicolumn{1}{c|}{\multirow{3}{*}{\textbf{Content-based}}} &\multicolumn{1}{c|}{\multirow{3}{*}{99.38\%}}&\multicolumn{1}{c|}{\multirow{3}{*}{91.31\%}}& 10\% &89.27\% / 1.71\% & 89.41\% / 3.66\% & 88.61\% / 12.74\% &90.31\% / 97.42\%&88.71\% / 47.94\%  &86.16\% / 2.78\%\\
\multicolumn{1}{c|}{} & \multicolumn{1}{c|}{} & \multicolumn{1}{c|}{} & 20\% & 90.60\% / 96.36\% &89.11\% / 37.27\% &87.02\% / 3.81\% & 89.44\% / 6.64\% & 89.81\% / 8.04\% & 88.69\% / 8.51\%  \\
\multicolumn{1}{c|}{} & \multicolumn{1}{c|}{} & \multicolumn{1}{c|}{} & 40\% &90.4\% / 92.68\% &89.76\% / 24.88\% &88.57\% / 1.54\% & 90.78\% / 6.11\% & 90.87\% / 9.23\% &90.36\% / 7.41\% \\\hline
\multicolumn{1}{c|}{\multirow{3}{*}{\textbf{Noise-based}}} &\multicolumn{1}{c|}{\multirow{3}{*}{100.00\%}}&\multicolumn{1}{c|}{\multirow{3}{*}{90.64\%}}& 10\% & 89.97\% / 84.43\% &88.74\% / 12.98\% &86.53\% / 3.79\% & 89.10\% / 1.17\% & 89.58\% / 0.37\% & 88.96\% / 1.42\% \\
\multicolumn{1}{c|}{} & \multicolumn{1}{c|}{} & \multicolumn{1}{c|}{} & 20\% &90.39\% / 82.67\% &89.23\% / 8.86\% &87.84\% / 1.86\% & 90.14\%  / 0.81\% &90.46\% / 1.39\% &89.86\%  / 0.92\%  \\
\multicolumn{1}{c|}{} & \multicolumn{1}{c|}{} & \multicolumn{1}{c|}{} & 40\% &90.47\% 87.83\% &90.12\% / 10.34\% &88.45\% / 2.41\% & 90.80\% / 0.59\% &90.62\% / 0.56\% & 90.42\% / 0.68\%   \\\hline
\multicolumn{1}{c|}{\multirow{3}{*}{\textbf{Unrelated}}} &\multicolumn{1}{c|}{\multirow{3}{*}{ 100.00\%}}&\multicolumn{1}{c|}{\multirow{3}{*}{91.69\%}}& 10\% &88.97\% / 100.00\% &87.72\% / 68.00\% &86.49\% / 41.00\% &87.12\% / 15.00\% &87.32\%  / 18.00\% &86.60\% / 22.00\%  \\
\multicolumn{1}{c|}{} & \multicolumn{1}{c|}{} & \multicolumn{1}{c|}{} & 20\% & 89.85\% / 100.00\%& 88.74\% / 35.00\% &87.53\% / 22.00\% & 89.06\% / 14.00\% &88.93\% / 16.00\%  &88.51\% / 17.00\%  \\
\multicolumn{1}{c|}{} & \multicolumn{1}{c|}{} & \multicolumn{1}{c|}{} & 40\% &90.61\% / 98.00\% &89.16\% / 31.00\% &88.01\% / 26.00\% &  89.55\% / 17.00\%& 89.67\% / 12.00\% &88.77\% / 16.00\% \\

\bottomrule[1.5pt]
\end{tabular}
}
\end{table*}

\subsection{Results}
The performance of watermark removal is assessed by two metrics: test accuracy and watermark retention. On the one hand, we observe the test accuracy on clean test set after watermark removal, on the other hand, we measure the proportion the watermarked inputs that are still misclassified into pre-defined labels. 


To illustrate that our approaches can successfully remove the watermark while preserving the test performance on original task, we first show the curves of watermark retention and test accuracy in Figure~\ref{fig:curves} as an example when we only have access to 20\% of the training data. At the beginning, the watermark retentions are almost 100\%, which means that the watermarked model memorizes the watermarks well. During the process of the retraining for watermark removal, the watermark retentions drop quickly  and are reduced to 
below  10\% for both content-based and noise-based watermarks. However, it is relatively more difficult to remove the unrelated watermarks and the watermark retention rates remain around 20\%.
The main reason is that the poisonous data used in unrelated watermark scheme comes from totally different domains. These various kinds of poisonous data greatly increase the difficulty of watermark removal.

Then we present the detailed results on test accuracy and watermark retention before and after watermark removal respectively. The results on MNIST are shown in Table~\ref{table:mnist}. It is obvious that CE and JS approaches have less impact on test accuracy compared with DN approach. When we only have 10\% of the training data, the test accuracy drops about 0.4\% for content-based watermarks and even slightly increases for noise-based watermarks by use of CE and JS. The impact on test accuracy is a little more for unrelated watermarks, with a drop around 1.3\%. In addition, unrelated watermarks are much more difficult to remove, with  higher watermark retentions after removal 
compared with the others. Besides, the more data we have access to, the higher test accuracy we get. When using 40\% of the training data, we hardly sacrifice the test accuracy while removing the watermarks. 

Also, as shown in Table~\ref{table:mnist}, REFIT cannot effectively remove the watermarks especially for noise-based and unrelated watermarks. It heavily relies on the learning rate settings of the optimization process, where an inappropriate learning rate leads to either high watermark retention or low test accuracy after fine-tuning. With the effect of EWC~\cite{kirkpatrick2017overcoming}, the test accuracy remains relatively high. However, compared to our result, the watermark removal ability cannot be guaranteed. In addition, we find that the watermark retention degrades slowly comparing to our method, which might further increase the computation power.


The detailed results on CIFAR-10 are present in Table~\ref{table:cifar}. It is clear that JS has the least impact on the test accuracy compared with the other two approaches. When we have access to 40\% of the training  data, the test accuracy drops only  0.5\% for content-based watermarks and hardly drop for noise-based scheme. Increasing the amount of available data mitigates the impact on test accuracy, and it is more evident in unrelated watermarks. This elaborates that as long as we have enough data, our approach will cause little degradation of the model performance. Similar to the results on MNIST, removing unrelated watermarks is relatively harder, with the watermark retention around 16\%. We can see taht all three approaches can effectively remove the watermark, however, CE and JS perform better than DN from the perspective of the impact on test accuracy.

Similarly, we find that REFIT~\cite{chen2019refit:} requires a large learning rate to remove the watermarks. However, the increase of learning rate could cause a decrease in test accuracy. Thus, REFIT~\cite{chen2019refit:} might require some extra information of the infused watermarks to achieve a good result in both hands. Combining all these results, our framework can effectively remove watermarks while preserving the performance on the original task.

\section{Conclusion}
In this work, we focus on the backdoor-based watermarking technology, which is widely applied to protect the intellectual property of the model. We demonstrate the  vulnerability of watermarking schemes and propose our framework \name{} for watermark removal. We aim to invalidate the watermark patterns so that the watermarked inputs can not be classified into pre-defined label. Specifically, our framework incorporates data augmentation and the optimization of distribution distance. In particular, our approach only requires limited data rather than the entire training set which is practical for adversaries in real situations. Compared with existing work, our approach does not require a carefully-designed learning rate schedule or additional unlabeled dataset. We evaluate the 
performance of our approach against three 
typical types of backdoor-based watermarks comparing to the state-of-the-art method. The results demonstrate that our approach can remove the watermark effectively with limited impact on the performance of the model.

\input{main.bbl}

\bibliographystyle{IEEEtran}

\end{document}